\documentclass{article}

\usepackage[final]{nips_2017}

\usepackage[utf8]{inputenc} 
\usepackage[T1]{fontenc}    
\usepackage{hyperref}       
\usepackage{url}            
\usepackage{booktabs}       
\usepackage{amsfonts}       
\usepackage{nicefrac}       
\usepackage{microtype}      
\usepackage{algorithm}
\usepackage{algorithmic}
\usepackage{amsmath}

\usepackage{enumitem}
\usepackage{float}
\usepackage{wrapfig}
\usepackage[page]{appendix}
\usepackage{mathrsfs} 
\usepackage{times}
\usepackage{graphicx} 
\usepackage{subfigure}
\usepackage{natbib}
\usepackage{algorithm}
\usepackage{algorithmic}
\usepackage{hyperref}
\usepackage{comment}
\usepackage{amsthm}
\usepackage{mathtools}
\usepackage{epstopdf}
\usepackage[colorinlistoftodos,prependcaption]{todonotes}
\newtheorem{theorem}{Theorem}
\newtheorem{lemma}{Lemma}

\newtheorem{definition}{Definition}
\newtheorem{example}{Example}

\newtheorem{assumption}{Assumption}

\title{Learning Causal Structures Using Regression Invariance}

\author{
  AmirEmad Ghassami$^{*\dagger}$, ~Saber Salehkaleybar$^{\dagger}$, ~Negar Kiyavash$^{*\dagger}$, ~Kun Zhang$^{\ddagger}$\\
  $^{*}$Department of ECE, University of Illinois at Urbana-Champaign, Urbana, USA.\\
  $^{\dagger}$Coordinated Science Laboratory, University of Illinois at Urbana-Champaign, Urbana, USA.\\
  $^{\ddagger}$Department of Philosophy, Carnegie Mellon University, Pittsburgh, USA.\\
  $^{\dagger}$\texttt{\{ghassam2,sabersk,kiyavash\}@illinois.edu}, $^{\ddagger}$\texttt{kunz1@cmu.edu} \\
}

\begin{document}

\maketitle

\begin{abstract}

We study causal inference in a multi-environment setting, in which the functional relations for producing the variables from their direct causes remain the same across environments, while the distribution of exogenous noises may vary.
We introduce the idea of using the invariance of the functional relations of the variables to their causes across a set of environments. We define a notion of completeness for a causal inference algorithm in this setting and prove the existence of such algorithm by proposing the baseline algorithm. Additionally, we present an alternate algorithm that has significantly improved computational and sample complexity compared to the baseline algorithm. The experiment results show that the proposed algorithm outperforms the other existing algorithms.

\end{abstract}


\vspace{-5mm}
\section{Introduction}
\label{sec:intro}
\vspace{-2mm}
Causal inference is a fundamental problem  in machine learning with applications in several fields such as biology, economics, epidemiology, computer science, etc. 
When performing interventions in the system is not possible (observation-only setting), the main approach to identifying direction of influences and learning the causal structure is to perform statistical tests based on the conditional dependency of the variables on the data \cite{pearl2009causality,spirtes2000causation}. In this case, a ``complete'' conditional independence based algorithm allows learning the causal structure to the extent possible, where by complete we mean that the algorithm is capable of distinguishing all the orientations up to the Markov equivalence. Such algorithms perform a conditional independence test along with the Meek rules\footnote{Recursive application of Meek rules identifies the orientation of additional edges to obtain the Markov equivalence class.
} introduced in \cite{verma1992algorithm}. IC \cite{judea1991equivalence} and PC \cite{spirtes1991algorithm} algorithms are two well known examples. Within the framework of structural equation models (SEMs) \cite{pearl2009causality}, by adding assumptions to the model such as non-Gaussianity \cite{shimizu2006linear}, nonlinearity \cite{hoyer2009nonlinear, peters2014causal} or equal noise variances \cite{peters2012identifiability}, it is even possible to identify the exact causal structure.
When the experimenter is capable of intervening in the system to see the effect of varying one variable on the other variables in the system (interventional setting), the causal structure could be exactly learned. In this setting, the most common identification procedure assumes that the variables whose distributions have varied are the descendants of the intervened variable and hence the causal structure is reconstructed by performing interventions in different variables in the system \cite{eberhardt2007causation,hauser2014two}.

\begin{wrapfigure}{r}{0.4\textwidth}
\vspace{-5mm}
  \begin{center}
    \includegraphics[width=0.4\textwidth]{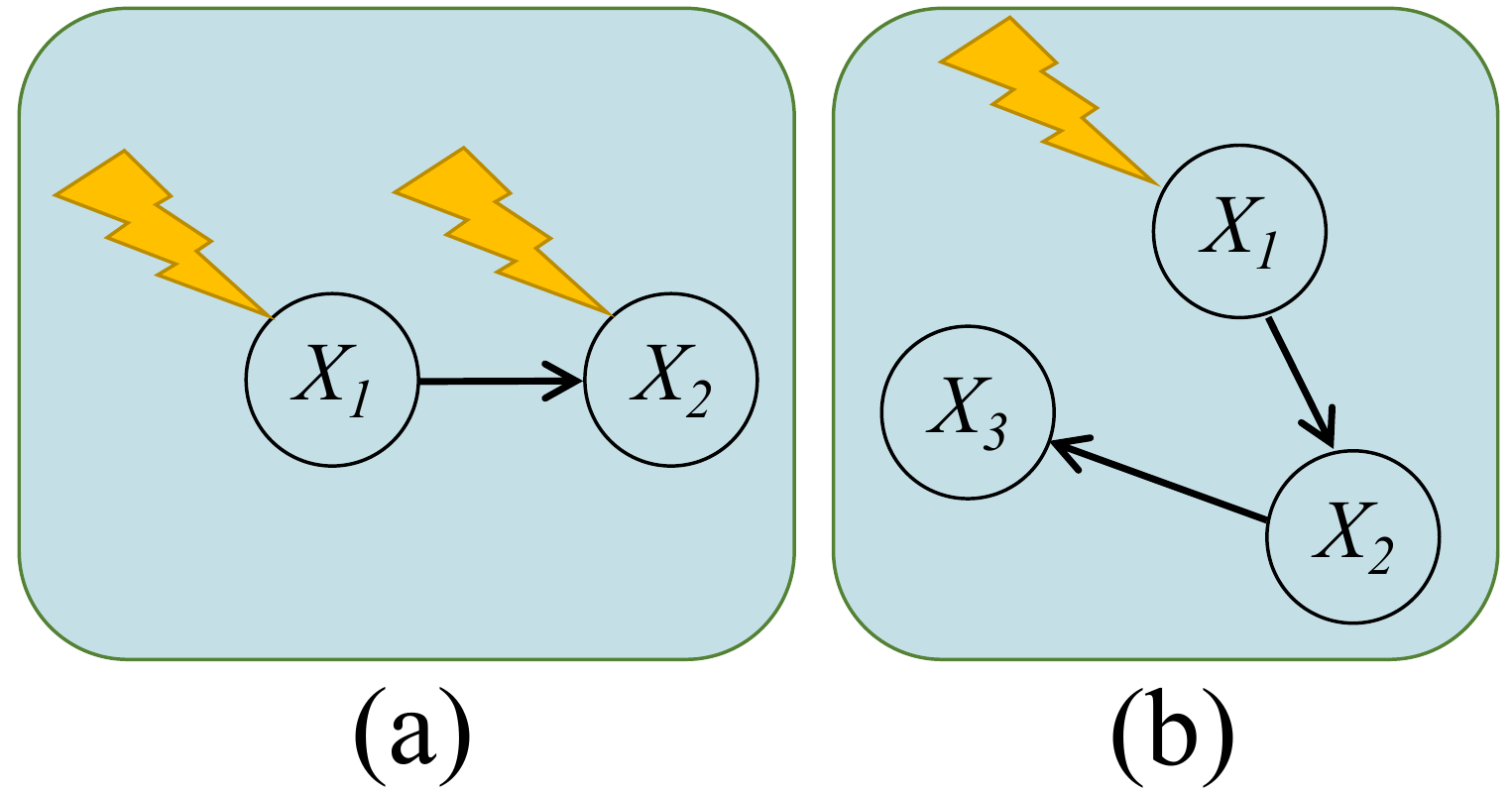}
  \end{center}
  \caption{Simple examples of identifiable structures using the proposed approach.}
  \label{fig:ex1}
  \vspace{-3mm}
\end{wrapfigure}

We take a different approach from the traditional interventional setting by considering a multi-environment setting, in which the functional relations for producing the variables from their parents remain the same across environments, while the distribution of exogenous noises may vary. This is different from the customary interventional setting, because in our model, the experimenter does not have any control on the location of the changes in the system, and as will be seen in Figure \ref{fig:ex1}(a), this may prevent the ordinary interventional approaches from working. The multi-environment setting was also studied in \cite{peters2016causal} and \cite{tian2001causal}; we will put our work into perspective in relationship to these in the related work below.

We focus on the linear SEM with additive noise as the underlying data generating model (see Section \ref{sec:desc} for details). Note that this model is one of the most problematic models in the literature of causal inference, and if the noises have Gaussian distribution, for many structures, none of the existing observational approaches can identify the underlying causal structure uniquely\footnote{As noted in \cite{hoyer2009nonlinear}, ``nonlinearities can play a role similar to that of non-Gaussianity'', and both lead to exact structure recovery.}. 
The main idea in our proposed approach is to utilize the change of the regression coefficients, resulting from the changes across the environments to distinguish causes from the effects. 

Our approach is able to identify causal structures previously not identifiable using the existing approaches. 
Figure \ref{fig:ex1} shows two simple examples to illustrate this point. In this figure, a directed edge form variable $X_i$ to $X_j$ implies that $X_i$ is a direct cause of $X_j$, and change of an exogenous noise across environments is denoted by the flash sign.
Consider the structure in  Figure \ref{fig:ex1}(a), with equations $X_1=N_1$, and $X_2=aX_1+N_2$, where $N_1\sim\mathcal{N}(0,\sigma_1^2)$ and $N_2\sim\mathcal{N}(0,\sigma_2^2)$ are independent mean-zero Gaussian exogenous noises. Suppose we are interested in finding out which variable is the cause and which is the effect. We are given two environments across which the exogenous noise of both $X_1$ and $X_2$ are varied. 
Denoting the regression coefficient resulting from regressing $X_i$ on $X_j$ by $\beta_{X_j}(X_i)$, in this case, we have
$
\beta_{X_2}(X_1)=\frac{Cov(X_1X_2)}{Cov(X_2)}=\frac{a\sigma_1^2}{a^2\sigma_1^2+\sigma_2^2},
$
and 
$
\beta_{X_1}(X_2)=\frac{Cov(X_1X_2)}{Cov(X_1)}=a.
$
Therefore, except for pathological cases for values for the variance of the exogenous noises in two environments, the regression coefficient resulting from regressing the cause variable on the effect variable varies between the two environments, while the regression coefficient from regressing the effect variable on the cause variable remains the same. Hence, the cause is distinguishable from the effect. Note that structures $X_1\rightarrow X_2$ and $X_2\rightarrow X_1$ are in the same Markov equivalence class and hence, not distinguishable using merely conditional independence tests. Also since the exogenous noises of both variables have changed, commonly used interventional tests are also not capable of distinguishing between these two structures \cite{eberhardt2005number}. Moreover, as it will be shortly explained (see related work), because the exogenous noise of the target variable has changed, the invariant prediction method \cite{peters2016causal}, cannot discern the correct structure either. \\
As another example, consider the structure in Figure \ref{fig:ex1}(b). Suppose the exogenous noise of $X_1$ is varied across the two environments. Similar to the previous example, it can be shown that $\beta_{X_2}(X_1)$ varies across the two environments while $\beta_{X_1}(X_2)$ remains the same. This implies that the edge between $X_1$ and $X_2$ is from the former to the later. Similarly, $\beta_{X_3}(X_2)$ varies across the two environments while $\beta_{X_2}(X_3)$ remains the same. This implies that $X_2$ is the parent of $X_3$. Therefore, the structure in Figure \ref{fig:ex1}(b) is distinguishable using the proposed identification approach. Note that the invariant prediction method cannot identify the relation between $X_2$ and $X_3$, and conditional independence tests are also not able to distinguish this structure.

\textbf{Related Work.} 
The best known algorithms for causal inference in the observational setup are IC \cite{judea1991equivalence} and PC \cite{spirtes1991algorithm} algorithms.
Such purely observational approaches reconstruct the causal graph up to Markov equivalence classes. Thus, directions of some edges may remain unresolved.
There are studies which attempt to identify the exact causal structure by restricting the model class \cite{shimizu2006linear, hoyer2009nonlinear, peters2012identifiability, ghassami2017interaction}. Most of such work consider SEM with independent noise. LiNGAM method \cite{shimizu2006linear} is a potent approach capable of structure learning in linear SEM model with additive noise\footnote{ There are extensions to LiNGAM beyond linear model \cite{zhang2008distinguishing}.}, as long as the distribution of the noise is not Gaussian. Authors of \cite{hoyer2009nonlinear} showed that nonlinearities can play a role similar to that of non-Gaussianity. 
In interventional approach for causal structure learning, the experimenter picks specific variables and attempts to learn their relation with other variables, by observing the effect of perturbing that variables on the distribution of others.
In recent work, bounds on the required number of interventions for complete discovery of causal relationships as well as passive and adaptive algorithms for  minimize the number of experiments were derived \cite{eberhardt2005number}  \cite{shanmugam2015learning}  \cite{hauser2014two} \cite{ghassami2017optimal}.\\
 In this work we assume that the functional relations of the variables to their direct causes across a set of environments are invariant. Similar assumptions have been considered in other work \cite{daniusis2012inferring,sgouritsa2015inference,janzing2010causal,janzing2012information,peters2016causal}. Specifically, \cite{daniusis2012inferring} which studies finding causal relation between two variables related to each other by an invertible function, assumes that `` the distribution of the cause and the function mapping cause to effect are independent since they correspond to independent mechanisms of nature''.\\
There is little work on multi-environment setup \cite{tian2001causal,peters2016causal,zhang2015discovery}.
In \cite{tian2001causal}, the authors analyze the classes of structures that are equivalent relative to a stream of distributions and present algorithms that output graphical representations of these equivalence classes. They assume that changing the distribution of a variable, varies the marginal distribution of all its descendants.
Naturally this also assumes that they have access to enough samples to test each variable for marginal distribution change. This approach cannot identify the causal relations among variables which are affected by environment changes in the same way.
The most closely related work to our approach is the invariant prediction method \cite{peters2016causal}, which utilizes different environments to estimate the set of predictors of a target variable.
In that work, it is assumed that the exogenous noise of the target variable does not vary among the environments. In fact, the method crucially relies on this assumption as it adds variables to the estimated predictors set only if they are necessary to keep the distribution of the target variable's noise fixed.
Besides high computational complexity, invariant prediction framework may result in  a set which does not contain all the parents of the target variable. Additionally, the optimal predictor set (output of the algorithm) is not necessarily unique. We will show that in many cases our proposed approach can overcome both these issues.
Recently, the authors of \cite{zhang2015discovery} considered the setting in which changes in the mechanism of variables prevents ordinary conditional independence based algorithms from discovering the correct structure. The authors have modeled these changes as multiple environments and proposed a general solution for a non-parametric model which first detects the variables whose mechanism changed and then finds causal relations among variables using conditional independence tests. Due to the generality of the model, this method requires a high number of samples. 

\textbf{Contribution.}
We propose a novel causal structure learning framework, which is capable of uniquely identifying structures which were not identifiable using existing methods. The main contribution of this work is to introduce the idea of using the invariance of the functional relations of the variables to their direct causes across a set of environments. This would imply the invariance of coefficients in the special case of linear SEM, in distinguishing the causes from the effects. We define a notion of completeness for a causal inference algorithm in this setting and prove the existence of such algorithm by proposing the baseline algorithm (Section 3). This algorithm first finds the set of variables for which distributions of 
noises have varied across the two environments, and then uses this information to identify the causal structure. Additionally, we present an alternate algorithm (Section \ref{sec:alg2}) which has significantly improved computational and sample complexity compared to the baseline algorithm.
\vspace{-3mm}
\section{Regression-Based Causal Structure Learning}
\label{sec:desc}
\vspace{-3mm}
\begin{definition}
Consider a directed graph $G=(V,E)$ with vertex set $V$ and set of directed edges $E$.
$G$ is a DAG if it is a finite graph with no directed cycles.
A DAG $G$ is called causal if its vertices represent random variables $V=\{X_1, ...,X_n\}$ and a directed edges $(X_i,X_j)$ indicates that variable $X_i$ is a direct cause of variable $X_j$.
\end{definition}
\vspace{-2mm}
We consider a linear SEM \cite{kenneth1989structural} 
as the underlying data generating model. In such a model
the value of each variable $X_j \in V$ is determined by a linear combination of the values of its causal parents $\textit{PA}(X_j)$ plus an additive exogenous noise $N_j$, where $N_j$'s are jointly independent as follows
\vspace{-1.5mm}
\begin{equation}
X_j = \sum_{X_i\in \textit{PA}(X_j)}b_{ji}X_i+N_j,\hspace{1cm}\forall  j\in \{1,\cdots,p\},
\label{eq:linear model}
\end{equation}
which could be represented by a single matrix equation $\mathbf{X}=\mathbf{B}\mathbf{X}+\mathbf{N}$. Further, we can write
\vspace{-1.5mm}
\begin{equation}
 \mathbf{X}=\mathbf{A}\mathbf{N},
 \label{eq:noise}
\end{equation} 
 where $\mathbf{A}=(\mathbf{I}-\mathbf{B})^{-1}$. 
This implies that each variable $X\in V$ can be written as a linear combination of the exogenous noises in the system. We assume that in our model, all variables are observable. Also, for the ease of representation, we focus on zero-mean Gaussian exogenous noise; otherwise, the results could be easily extended to any arbitrary distribution for the exogenous noise in the system. The following definitions will be used throughout the paper.
\begin{definition}
Graph union of a set $\mathcal{G}$ of mixed graphs\footnote{A mixed graph contains both directed and undirected edges.} over a skeleton, is a mixed graph with the same skeleton as the members of $\mathcal{G}$ which contains directed edge $(X,Y)$, if $\exists ~G\in\mathcal{G}$ such that $(X,Y)\in E(G)$ and $\not\exists ~G'\in\mathcal{G}$ such that $(Y,X)\in E(G')$. The rest of the edges 
remain undirected.
\end{definition}

\begin{definition}
Causal DAGs $G_1$ and $G_2$ over $V$ are Markov equivalent if every distribution that is compatible with one of the graphs is also compatible with the other. Markov equivalence is an equivalence relationship over the set of all graphs over $V$ \cite{koller2009probabilistic}. 
The graph union of all DAGs in the Markov equivalence class of a DAG $G$ is called the essential graph of $G$ and is denoted by $\textit{Ess}(G)$.
\end{definition} 
\vspace{-1mm}

We consider a multi-environment setting consisting of $M$ environments $\mathcal{E}=\{E_1,...,E_M\}$. The structure of the causal DAG and the functional relations for producing the variables from their parents (the matrix $\mathbf{B}$), remains the same across all environments, the exogenous noises may vary though.  
For a pair of environments $E_i,E_j\in\mathcal{E}$, let $I_{ij}$ be the set of variables whose exogenous noise changed between the two environments.
Given $I_{ij}$, for any DAG $G$ consistent with the essential graph\footnote{DAG $G$ is consistent with mixed graph $M$, if $G$ does not contain edge $(X,Y)$ while $M$ contains $(Y,X)$.} obtained from the conditional independence test, define the regression invariance set as follows
\vspace{-1mm}
\[
R(G,I_{ij})\coloneqq\{(X,S):X\in V,S\subseteq V\backslash\{X\},\beta^{(i)}_S(X)=\beta^{(j)}_S(X)\},
\]
where $\beta^{(i)}_S(X)$ and $\beta^{(j)}_S(X)$ are the regression coefficients of regressing variable $X$ on $S$ in environments $E_i$ and $E_j$, respectively. In words, for all variables $X\in V$, $R(G,I_{ij})$ contains all subsets $S\subseteq V\backslash\{X\}$ that if we regress $X$ on $S$, the regression coefficients do not change across $E_i$ and $E_j$.
\begin{definition}
\label{def:indist}
Given $I$, the set of variables whose exogenous noise has changed between two environments, DAGs $G_1$ and $G_2$ are called $I$-distinguishable if $R(G_1,I)\neq R(G_2,I)$.
\end{definition}
%
%
\vspace{-2mm}

We make the following assumption on the distributions of the exogenous noises. The purpose of this assumption is to rule out pathological cases for values of the variance of the exogenous noises in two environments which make special regression relations. For instance, in Example \ref{fig:ex1}, $\beta_{X_2}^{(1)} (X_1)=\beta_{X_2}^{(2)}(X_1)$ only if $\sigma_1^2 \tilde{\sigma}_2^2=\sigma_2^2 \tilde{\sigma}_1^2$ where $\sigma_i^2$ and $\tilde{\sigma}_i^2$ are the variances of the exogenous noise of $X_i$ in the environments $E_1$ and $E_2$, respectively. Note that this special relation between $\sigma_1^2$, $\tilde{\sigma}_1^2$, $\sigma_2^2$, and $\tilde{\sigma}_2^2$ has Lebesgue measure zero in the set of all possible values for the variances. 
\begin{assumption}[Regression Stability Assumption]
\label{ass:stability}
For a given set $I$ and structure $G$, perturbing the variance of the distributions of the exogenous noises by a small value $\epsilon$ does not change the regression invariance set $R(G,I)$.
\end{assumption}
\vspace{-3mm}


We give the following examples as applications of our approach.
\begin{example}
Consider DAGs $G_1:X_1\rightarrow X_2$ and $G_2:X_1\leftarrow X_2$.
For $I=\{X_1\}$, $I=\{X_2\}$ or $I=\{X_1,X_2\}$, calculating the regression coefficients as explained in Section \ref{sec:intro}, we see that $(X_1,\{X_2\})\not\in R(G_1,I)$ but $(X_1,\{X_2\})\in R(G_2,I)$. Hence $G_1$ and $G_2$ are $I$-distinguishable.
As mentioned in Section \ref{sec:intro}, structures $G_1$ and $G_2$ are not distinguishable using the ordinary conditional independence tests. Also, in the case of $I=\{X_1,X_2\}$, the invariant prediction approach and the ordinary interventional tests - in which the experimenter expects that a change in the distribution of the effect would not perturb the marginal distribution of the cause variable - are not capable of distinguishing the two structures either.
\end{example}
\vspace{-2mm}
~\vspace{-10mm}
\begin{wrapfigure}{r}{0.4\textwidth}
  \begin{center}
    \includegraphics[width=0.4\textwidth]{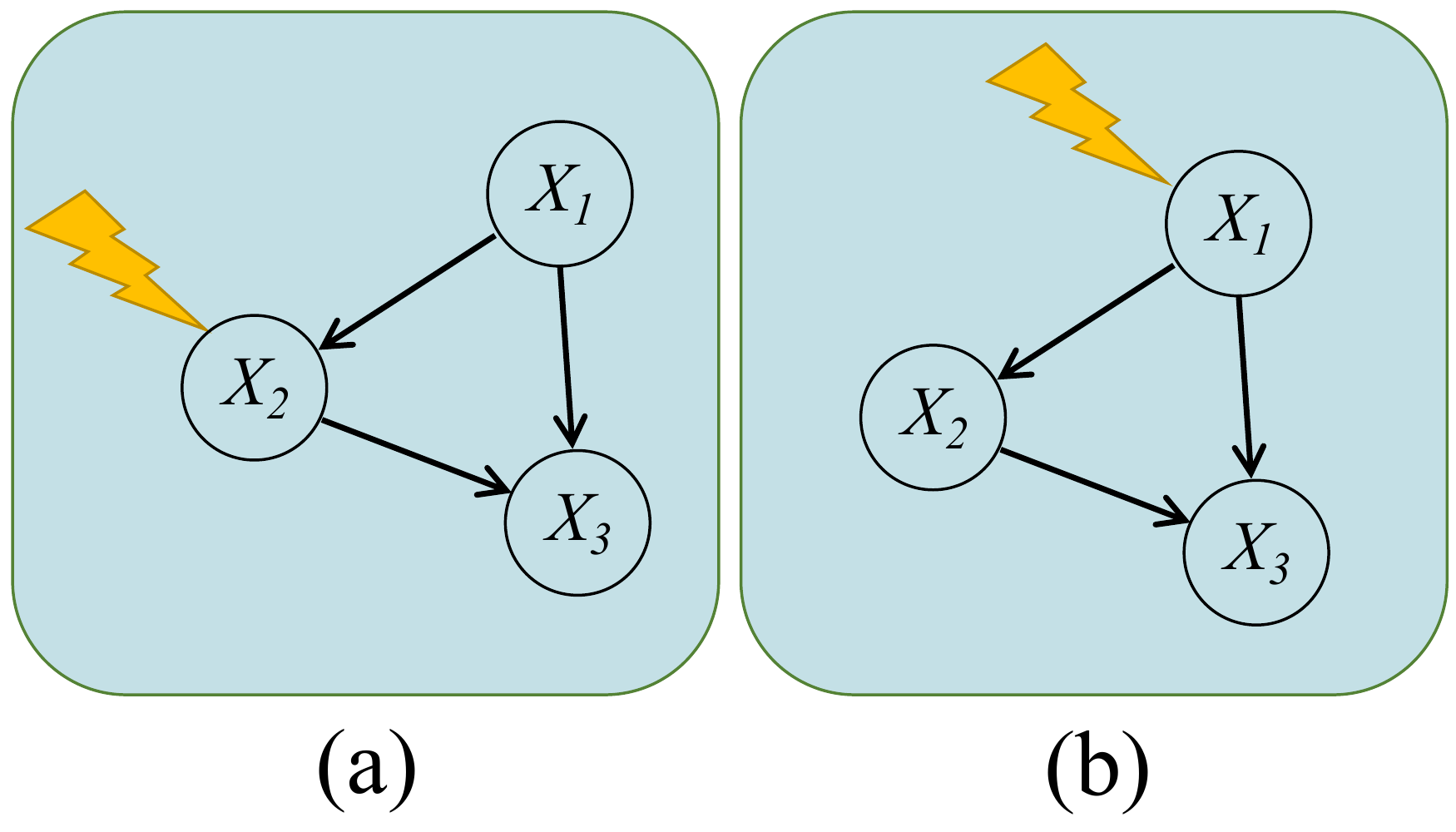}
  \end{center}
  \caption{DAG related to Example 3.}
  \label{fig:ex2}
  \vspace{-4mm}
\end{wrapfigure}
\begin{example}
Consider the DAG $G$ in Figure \ref{fig:ex1}$(b)$ with $I=\{X_1\}$. Consider an alternative DAG $G'$ in which compared to $G$ the directed edge $(X_1,X_2)$ is replaced by $(X_2,X_1)$, and DAG $G''$ in which compared to $G$ the directed edge $(X_2,X_3)$ is replaced by $(X_3,X_2)$. Since $(X_2,\{X_1\})\in R(G,I)$ while this pair is not in $R(G',I)$, and $(X_2,\{X_3\})\not\in R(G,I)$ while this pair belongs to $R(G'',I)$, the structure of $G$ is also distinguishable using the proposed identification approach. Note that $G$ is not distinguishable using conditional independence tests. Also, the invariant prediction method cannot identify the relation between $X_2$ and $X_3$, since it can keep the variance of the noise of $X_3$   fixed by setting the predictor set as $\{X_2\}$ or $\{X_1\}$, which have empty intersection.
\vspace{-1mm}
\end{example}
\begin{example}
\label{ex:tri}
Consider the structure in Figure \ref{fig:ex2}$(a)$ with $I=\{X_2\}$. Among the six possible triangle DAGs, all of them are $I$-distinguishable from this structure and hence, with two environments differing in the exogenous noise of $X_2$, this triangle DAG could be identified. Note that all the triangle DAGs are in the same Markov equivalent class and hence, using the information of one environment alone, observation only setting cannot lead to identification.
For $I=\{X_1\}$, the structure in Figure \ref{fig:ex2}$(b)$ is not $I$-distinguishable from a triangle DAG in which the direction of the edge $(X_2,X_3)$ is flipped. These two DAGs are also not distinguishable using usual intervention analysis and the invariant prediction method. 
\end{example}
\vspace{-2mm}
Let the structure $G^*$ be the ground truth DAG structure.
Define $\mathcal{G}_I\coloneqq\{G:R(G,I)=R(G^*,I)\}$, which is the set of all DAGs which are not $I$-distinguishable from $G^*$.
Using this set, we form the mixed graph $M_I$ over $V$, as the graph union of members of $\mathcal{G}_I$.
\begin{definition}
\label{def:completeness}
An algorithm $\mathscr{A}:(\textit{Ess}(G),R)\rightarrow M$ which gets an essential graph and a regression invariance set as the input and returns a mixed graph, is regression invariance complete if 
$
\mathscr{A}(\textit{Ess}(G^*),R(G^*,I))=M_I.
$
for any directed graph $G^*$ and set $I$. 
\end{definition}
\vspace{-3mm}
In other words, we say an algorithm $\mathscr{A}$ is regression invariance complete if given the correct essential graph and regression invariance set, it is able to return the appropriate  mixed graph. 
In Section \ref{sec:alg1} we will introduce a structure learning algorithm which is complete in the sense of Definition \ref{def:completeness}. 

\vspace{-3mm}
\section{Existence of Complete Algorithms}
\label{sec:alg1}
\vspace{-2mm}
In this section we show the existence of complete algorithm for learning the causal structure among a set of variables $V$ whose dynamics satisfy the SEM in \eqref{eq:linear model} in the sense of Definition \ref{def:completeness}. The pseudo-code of the algorithm is presented in Algorithm \ref{alg:basic}. 

\begin{wrapfigure}{r}{0.5\textwidth}
    \begin{minipage}{0.5\textwidth}
    \vspace{-7mm}
\begin{algorithm}[H]
\begin{algorithmic}
 \STATE {\bf Input:} Joint distribution over $V$ in environments $\mathcal{E}=\{E_i\}_{i=1}^M$.
 \STATE Obtain $\textit{Ess}(G^*)$ by performing a complete conditional independence test.
\FOR{each pair of environments $\{E_i,E_j\}\subseteq\mathcal{E}$}
\STATE Obtain $R_{ij}=\{(Y,S):Y\in V, S\subseteq V\backslash \{Y\}, \beta^{(i)}_S(Y)=\beta^{(j)}_S(Y)\}$.
\STATE $I_{ij}=ChangeFinder(E_i,E_j)$. 
\STATE $\mathcal{G}_{ij}=ConsistentFinder(\textit{Ess}(G^*),R_{ij},I_{ij})$.
\STATE $M_{ij}=\bigcup_{G\in\mathcal{G}_{ij}} G$.
\ENDFOR
\STATE $M_{\mathcal{E}}= \bigcup_{1\leq i,j\leq M} M_{ij}$.
\STATE Perform Meek rules on $M_{\mathcal{E}}$ to get $\hat{M}$.
\STATE {\bf Output:} Mixed graph $\hat{M}$.
 \caption{The Baseline Algorithm}
 \label{alg:basic}
\end{algorithmic}
\end{algorithm}
\vspace{-10mm}
  \end{minipage}
  \end{wrapfigure}

Suppose $G^*$ is the ground truth structure. The algorithm first performs a conditional independence test followed by applying Meek rules to obtain the essential graph $\textit{Ess}(G^*)$. For each pair of environments $\{E_i,E_j\}\in\mathcal{E}$, first the algorithm calculates the regression coefficients $\beta^{(i)}_S(Y)$ and $\beta^{(j)}_S(Y)$, for all $Y\in V$ and $S\subseteq V\backslash \{Y\}$, and forms the regression invariance set $R_{ij}$, which contains the pairs $(Y,S)$ for which the regression coefficients did not change between $E_i$ and $E_j$.
Next, using the function ChangeFinder$(\cdot)$, we discover the set $I_{ij}$ which is the set of variables whose exogenous noises have varied between the two environments $E_i$ and $E_j$.
Then using the function ConsistantFinder$(\cdot)$, we find $\mathcal{G}_{ij}$ which is the set of all possible DAGs, $G$ that is consistent with $\textit{Ess}(G^*)$ and $R(G,I_{ij})=R_{ij}$. After taking the union of graphs in $\mathcal{G}_{ij}$, we form the graph $M_{ij}$ which is the mixed graph containing all causal relations distinguishable from the given regression information between the two environments. Clearly, since we are searching over all DAGs, the baseline algorithm is complete in the sense of Definition \ref{def:completeness}.

After obtaining $M_{ij}$ for all pairs of environments, the algorithm forms a mixed graph $M_{\mathcal{E}}$ by taking graph union of $M_{ij}$'s. We perform the Meek rules on $M_{\mathcal{E}}$ to find all extra orientations and output $\hat{M}$.
%

\textbf{Obtaining the set $R_{ij}$:}
In this part, for a given significance level $\alpha$, we will show how the set $R_{ij}$ can be obtained correctly with probability at least $1-\alpha$.
For given $Y\in V$ and $S\subseteq V\backslash \{Y\}$ in the environments $E_i$ and $E_j$, we define the null hypothesis $H_{0,Y,S}^{ij}$ as follows:
\vspace{-1mm}
\begin{equation}
H_{0,Y,S}^{ij}: \exists \beta \in \mathbb{R}^{|S|} \mbox{ such that } \beta^{(i)}_S(Y)=\beta \mbox{ and }\beta^{(j)}_S(Y)=\beta.
\label{eq:HT}
\end{equation}
Let $\hat{\beta}^{(i)}_S(Y)$ and $\hat{\beta}^{(j)}_S(Y)$ be the estimations of $\beta^{(i)}_S(Y)$ and $\beta^{(j)}_S(Y)$, respectively, obtained using the ordinary least squares estimator computed from observational data.  If the null hypothesis is true, then 
\vspace{-1mm}
\begin{equation}
(\hat{\beta}^{(i)}_S(Y)-\hat{\beta}^{(j)}_S(Y))^T (s_i^2\Sigma_i^{-1}+s_j^2\Sigma_j^{-1})^{-1} (\hat{\beta}^{(i)}_S(Y)-\hat{\beta}^{(j)}_S(Y))/p \sim F(p,n-p),
\label{eq:Ftest}
\end{equation}
where $s^2_i$ and $s_j^2$ are unbiased estimates of variance of $Y^{(i)}-X_S^{(i)}\beta^{(i)}_S(Y)$ and $Y^{(j)}-X_S^{(j)}\beta^{(j)}_S(Y)$, respectively 
(see Appendix \ref{app:eq4} for details).
 Furthermore, we have $\Sigma_i=(X^{(i)}_S)^T X^{(i)}_S$ and $\Sigma_j=(X^{(j)}_S)^T X^{(j)}_S$. 

We reject the null hypothesis $H_{0,Y,S}^{ij}$ if the p-value of (\ref{eq:Ftest}) is less than $\alpha/(p\times (2^{p-1}-1))$. By testing all null hypotheses $H_{0,Y,S}^{ij}$ for any $Y\in V$ and $S\subseteq V\backslash \{Y\}$, we can obtain the set $R_{ij}$ correctly with probability at least $1-\alpha$.

\textbf{Function \textit{ChangeFinder}($\cdot$):}
We use Lemma \ref{lem:change} to find the set $I_{ij}$ with probability at least $1-2\alpha$.
\begin{lemma}
Given environments $E_i$ and $E_j$, for a variable $Y\in V$, if $\mathbb{E}\{(Y^{(i)}-X_S^{(i)}\beta^{(i)}_S(Y))^2\}\neq \mathbb{E}\{(Y^{(j)}-X_S^{(j)}\beta_S^{(j)}(Y))^2\}$ for all $S\subseteq N(Y)$ such that $(Y,S)\in R_{ij}$, where $N(Y)$ is the set of neighbors of $Y$, then the variance of exogenous noise $N_Y$ is changed between the two environments. Otherwise, the variance of $N_Y$ is fixed.
\label{lem:change}
\end{lemma}
\vspace{-2mm}
See Appendix \ref{app:lemchange} for the proof.


Based on Lemma \ref{lem:change}, we try to find a set $S\subseteq N(Y), (Y,S)\in R_{ij}$ such that the variance of residual $Y-X_S\beta_S(Y)$ remains fixed between two environments. To do so, we check whether the variance of exogenous noise $N_Y$ is changed between two environments $E_i$ and $E_j$ by testing the following null hypothesis for any set $S\subseteq N(Y), (Y,S)\in R_{ij}$:
$\bar{H}_{0,Y,S}^{ij}: \exists \sigma\in \mathbb{R} \mbox{ s.t. } \mathbb{E}\{(Y^{(i)}-X^{(i)}_S\beta_S^{(i)}(Y))^2\}=\sigma^2 \mbox{ and } \mathbb{E}\{(Y^{(j)}-X_{S}^{(j)}\beta_{S}^{(j)}(Y))^2\}=\sigma^2$.

In order to test the above null hypothesis, we can compute the variance of residuals $Y^{(i)}-X_{S}^{(i)}\hat{\beta}_{S}^{(i)}$ and $Y^{(j)}-X_{S}^{(j)}\hat{\beta}_{S}^{(j)}$ and test whether these variances are equal using an $F$-test. If the p-value for the set $S$ is less than $\alpha/(p\times (2^\Delta-1))$, then we will reject the null hypothesis $\bar{H}_{0,Y,S}^{ij}$ where $\Delta$ is the maximum degree of the causal graph. If  we reject all hypothesis tests $\bar{H}_{0,Y,S}^{ij}$ for any $S\in N(Y), (Y,S)\in R_{ij}$, then we will add $Y$ to set $I_{ij}$.

%
\textbf{Function \textit{ConsistentFinder}($\cdot$):}
Let $D_{st}$ be the set of all directed paths from variable $X_s$ to variable $X_t$. For any $d\in D_{st}$, we define the weight of directed path $d \in D_{st}$ as $w_d:= \Pi_{(u,v)\in d} b_{vu}$ where $b_{vu}$ are coefficients in \eqref{eq:linear model}. By this definition, it can be seen that the entry $(t,s)$ of matrix $\mathbf{A}$ in \eqref{eq:noise} is equal to $[\mathbf{A}]_{ts}=\sum_{d\in D_{st}} w_d$. Thus, the entries of matrix $\mathbf{A}$ are multivariate polynomials of entries of $\mathbf{B}$. Furthermore,
\begin{equation}
\beta_S^{(i)}(Y)= \mathbb{E} \{ X^{(i)}_S (X^{(i)}_S)^T \}^{-1} \mathbb{E} \{X^{(i)}_S Y^{(i)}\}= (\mathbf{A}_S \mathbf{\Lambda}_i \mathbf{A}_S^T)^{-1} \mathbf{A}_S \mathbf{\Lambda}_i \mathbf{A}^T_Y,
\label{eq:linear regression}
\end{equation}
where $\mathbf{A}_S$ and $\mathbf{A}_Y$ are the rows corresponding to set $S$ and $Y$ in matrix $\mathbf{A}$, respectively and matrix $\mathbf{\Lambda}_i$ is a diagonal matrix where $[\mathbf{\Lambda}_i]_{kk}= \mathbb{E}\{(N^{(i)}_k)^2\}$.  

From the above discussion, we know that the entries of matrix $\mathbf{A}$ are multivariate polynomials of entries of $\mathbf{B}$. Equation (\ref{eq:linear regression}) implies that the entries of vector $\beta_S^{(i)}(Y)$ are rational functions of entries in $\mathbf{B}$ and $\mathbf{\Lambda}_i$. Therefore, the entries of Jacobian matrix of $\beta_S^{(i)}(Y)$ with respect to the diagonal entries of $\mathbf{\Lambda}_i$ are also rational expression of these parameters. 

In function \textit{ConsistentFinder}(.), we select any directed graph $G$ consistent with  $Ess(G^*)$ and set $b_{vu}=0$ if $(u,v)\not\in G$. In order to check whether $G$ is in $\mathcal{G}_{ij}$, we initially set $R(G,I_{ij})=\emptyset$. Then, we compute the Jacobian matrix of $\beta_S^{(i)}(Y)$ parametrically for any $Y\in V$ and $S\in V\backslash \{Y\}$. As noted above, the entries of Jacobian matrix can be obtained as rational expressions of entries in $\mathbf{B}$ and $\mathbf{\Lambda}_i$. If all columns of Jacobian matrix corresponding to the elements of $I_{ij}$ are zero, then we add $(Y,S)$ to set $R(G,I_{ij})$ (since $\beta^{(i)}_S(Y)$ is not changing by varying the variances of exogenous noises in $I_{ij}$). After checking all $Y\in V$ and $S\in V\backslash \{Y\}$, we consider the graph $G$ in $\mathcal{G}_{ij}$ if $R(G,I_{ij})=R_{ij}$.


%

\vspace{-3mm}
\section{LRE Algorithm}
\label{sec:alg2}
\vspace{-2mm}
The baseline algorithm of Section \ref{sec:alg1} is presented to prove the existence of complete algorithms but it is not practical due to its high computational and sample complexity.
In this section we present the \textit{Local Regression Examiner} (LRE) algorithm, which is an alternative much more efficient algorithm for learning the causal structure among a set of variables $V$. The pseudo-code of the algorithm is presented in Algorithm \ref{alg:2}. We make use of the following result in this algorithm.
\begin{lemma}
\label{lem:fv}
Consider adjacent variables $X, Y\in V$ in causal structure $G$. For a pair of environments $E_i$ and $E_j$, if $(X,\{Y\})\in R(G,I_{ij})$, but $(Y,\{X\})\not\in R(G,I_{ij})$, then $X$ is the parent of $Y$.
\end{lemma}
\vspace{-2mm}
See Appendix \ref{app:lemfv} for the proof.
\vspace{-1mm}

%

\begin{algorithm}[t]
\begin{algorithmic}
 \STATE {\bf Input:} Joint distribution over $V$ in environments $\mathcal{E}=\{E_i\}_{i=1}^M$.
 \STATE {\bf Stage 1:} Obtain $\textit{Ess}(G^*)$ by performing a complete conditional independence test, and for all $X\in V$, form $\textit{PA}(X)$, $\textit{CH}(X)$, $\textit{UK}(X)$.
  \STATE {\bf Stage 2:} 
  \FOR{each pair of environments $\{E_i,E_j\}\subseteq\mathcal{E}$}
\FOR{all $Y\in V$}
\FOR{each $X\in \textit{UK}(Y)$}
\STATE Compute $\beta^{(i)}_X(Y)$, $\beta^{(j)}_{X}(Y)$, $\beta^{(i)}_Y(X)$, and $\beta^{(j)}_{Y}(X)$.
        \IF{$\beta^{(i)}_X(Y)\neq\beta^{(j)}_{X}(Y)$, but  $\beta^{(i)}_Y(X)=\beta^{(j)}_{Y}(X)$}
        	\STATE Set $X$ as a child of $Y$ and set $Y$ as a parent of $X$.
        \ELSIF{$\beta^{(i)}_X(Y)=\beta^{(j)}_{X}(Y)$, but  $\beta^{(i)}_Y(X)\neq\beta^{(j)}_{Y}(X)$}
        \STATE Set $X$ as a parent of $Y$ and set $Y$ as a child of $X$.
        \ELSIF{$\beta^{(i)}_X(Y)\neq\beta^{(j)}_{X}(Y)$, and  $\beta^{(i)}_Y(X)\neq\beta^{(j)}_{Y}(X)$}
        \STATE Find minimum set $S\subseteq N(Y)\backslash \{X\}$ such that $\beta^{(i)}_{S\cup \{X\}}(Y)=\beta^{(j)}_{S\cup \{X\}}(Y)$.
        \IF{$S$ does not exist }
        	\STATE Set $X$ as a child of $Y$ and set $Y$ as a parent of $X$.
        \ELSIF{$\beta^{(i)}_{S}(Y)\neq\beta^{(j)}_{S}(Y)$}
        \STATE $\forall W\in \{X\}\cup S$, set $W$ as a parent of $Y$ and set $Y$ as a child of $W$.
        \ELSE
        \STATE $\forall W\in S$, set $W$ as a parent of $Y$ and set $Y$ as a child of $W$.
        \ENDIF
        \ENDIF
\ENDFOR 
\ENDFOR 
\ENDFOR
\STATE {\bf Stage 3:} Perform Meek rules on the resulted mixed graph to obtain $\hat{M}$.
\STATE {\bf Output:} Mixed graph $\hat{M}$.
 \caption{LRE Algorithm}
 \label{alg:2}
\end{algorithmic}
\end{algorithm}

LRE algorithm consists of three stages. In the first stage, similar to the baseline algorithm, it performs a complete conditional independence test to obtain the essential graph. Then for each variable $X\in V$, it forms the set of $X$'s discovered parents, $\textit{PA}(X)$, and discovered children, $\textit{CH}(X)$, and leaves the remaining neighbors as unknown in $\textit{UK}(X)$.
In the second stage, the goal is that for each variable $Y\in V$, we find $Y$'s relation with its neighbors in $\textit{UK}(Y)$, based on the invariance of its regression on its neighbors across each pair of environments.
To do so, for each pair of environments, after fixing a target variable $Y$ and for each of its neighbors in $\textit{UK}(X)$, the regression coefficients of $X$ on $Y$ and $Y$ on $X$ are calculated. We will face one of the following cases:
\vspace{-2mm}
\begin{itemize}[leftmargin=5mm]
\item If neither is changing, we do not make any decisions about the relationship of $X$ and $Y$. This case is similar to having only one environment, similar to the setup in \cite{shimizu2006linear}.\vspace{-1mm}
\item If one is changing and the other is fixed, Lemma \ref{lem:fv} implies that the variable which fixes the coefficient as the regressor is the parent.\vspace{-1mm}
\item If both are changing, we look for an auxiliary set $S$ among $Y$'s neighbors with minimum number of elements, for which $\beta^{(i)}_{S\cup \{X\}}(Y)=\beta^{(j)}_{S\cup \{X\}}(Y)$. If no such $S$ is found, it implies that $X$ is a child of $Y$. Otherwise, if $S$ and $X$ are both required in the regressors set to fix the coefficient, 
we set $\{X\}\cup S$ as parents of $Y$; otherwise, if $X$ is not required in the regressors set to fix the coefficient, although we still set $S$ as parents of $Y$, we do not make any decisions regarding the relation of $X$ and $Y$ (Example \ref{ex:tri} when $I=\{X_1\}$, is an instance of this case).
\end{itemize}
\vspace{-2mm}
After adding the discovered relationships to the initial mixed graph, in the third stage, we perform the Meek rules on resulting mixed graph to find all extra possible orientations and output $\hat{M}$.

\textbf{Analysis of the Refined Algorithm.}
We can use the hypothesis testing in (\ref{eq:HT}) to test whether two vectors $\beta_S^{(i)}(Y)$ and $\beta_S^{(j)}(Y)$ are equal for any $Y\in V$ and $S\subseteq N(Y)$. If the p-value for the set $S$ is less than $\alpha/(p\times (2^{\Delta}-1))$, then we will reject the null hypothesis $H_{0,Y,S}^{ij}$. By doing so, the output of the algorithm will be correct with probability at least $1-\alpha$.
Regarding the computational complexity, since for each pair of environments, in the worse case we perform $\Delta (2^\Delta-1)$ hypothesis tests for each variable $Y\in V$, and considering that we have $\binom{M}{2}$ pairs of environments, the computational complexity of LRE algorithm is in the order of $\binom{M}{2} p\Delta(2^\Delta-1)$. Therefore, the bottleneck in the complexity of LRE is having to perform a complete conditional independence test in its first stage.
\begin{figure}
  \begin{center}
    \includegraphics[scale=0.255]{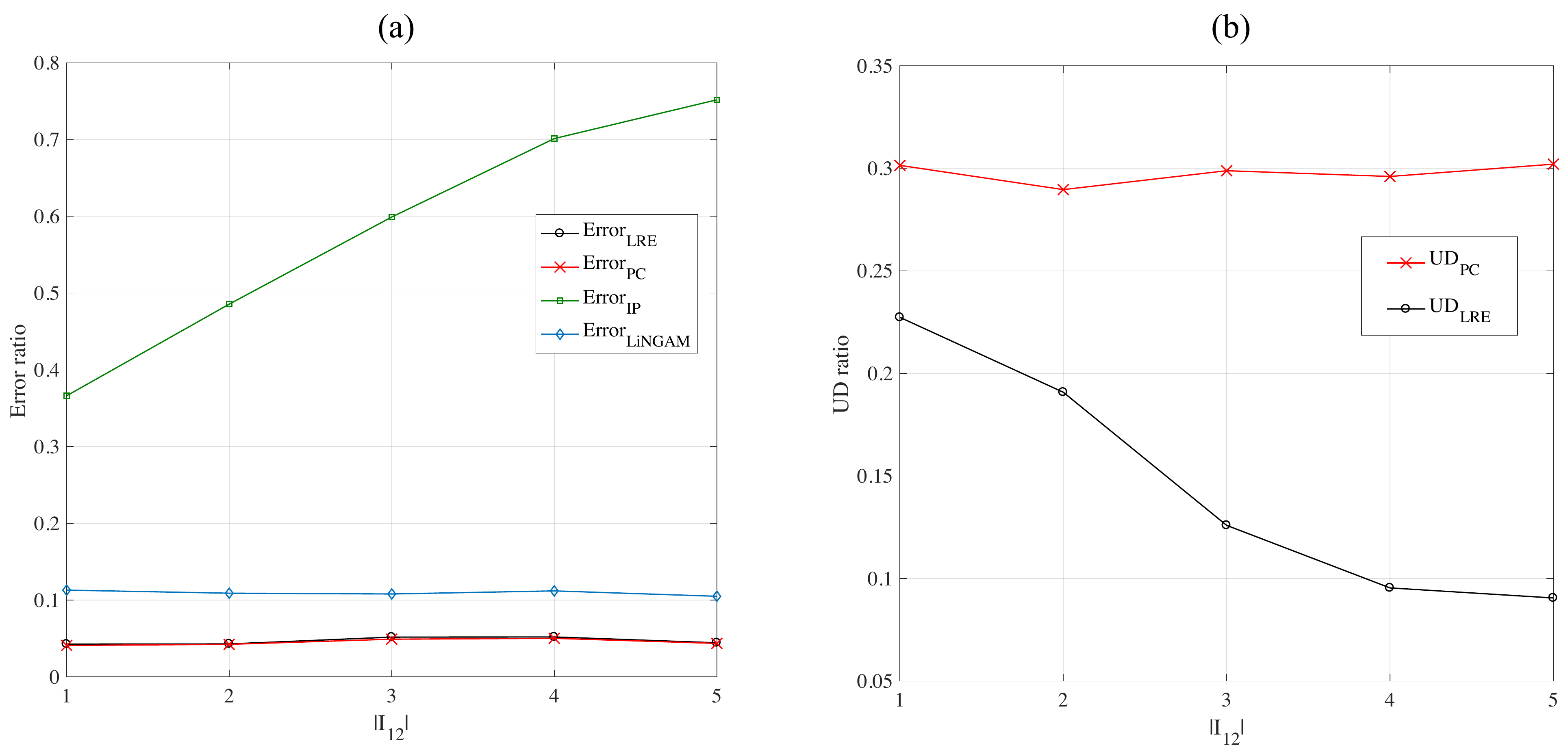}
  \end{center}
  \caption{(a) Error ration of LRE, PC and IP algorithms, (b) UD ratio of LRE and PC algorithms.}\vspace{-3mm}
  \label{fig:sim}
\end{figure}
\vspace{-3mm}
\section{Experiments}
\label{sec:exps}
\vspace{-2mm}
We evaluate the performance of LRE algorithm by testing it on both synthetic and real data.
As seen in the pseudo-code in Algorithm \ref{alg:2}, LRE has three stages where in the first stage, a complete conditional independence is performed. In order to have acceptable time complexity, in our simulations, we used the PC algorithm\footnote{We use the \textbf{pcalg} package \cite{kalisch2012causal} to run the PC algorithm on a set of random variables.} \cite{spirtes2000causation}, which is known to have a complexity of order $O(p^{\Delta})$ when applied to a  graph of order $p$ with degree bound $\Delta$.

\textbf{Synthetic Data.}
We generated 100 DAGs of order $p=10$ by first selecting a causal order for variables and then connecting each pair of variables with probability $0.25$. We generated data from a linear Gaussian SEM with coefficients drawn uniformly at random from $[0.1,2]$, and the variance of each exogenous noise was drawn uniformly at random from $[0.1,4]$. For each variable of each structure, $10^5$
  samples were generated.
In our simulation, we only consider a scenario in which we have two environments $E_1$ and $E_2$, where in the second environment, the exogenous noise of $|I_{12}|$ variables were varied. The perturbed variables were chosen uniformly at random.\\
Figure \ref{fig:sim} shows the error ratio and undirected edges (UD) ratio, for stage 1, which corresponds to the PC algorithm, and for the final output of LRE algorithm. Define a link to be any directed or undirected edge. The error ratio is calculated as follows:
$\textit{Error ratio}\coloneqq(|\textit{miss-detected links}|+|\textit{extra detected links}|+|\textit{wrongly oriented edges}|)/\binom{p}{2}$.
For the UD ratio, we count the number of undirected edges only among correctly detected links, i.e.,
$
\textit{UD ratio}\coloneqq(|\textit{correctly detected undirected edges}|)/(|\textit{correctly detected directed edges}|+|\textit{correctly detected undirected edges}|).
$
As seen in Figure \ref{fig:sim}, only one change in the second environment (i.e., $|I_{12}|=1$), reduces the UD ratio by $8$ percent compared to the PC algorithm. Also, the main source of error in LRE algorithm results from the application of the PC algorithm.
We also compared the error ratio of LRE algorithm with the Invariant Prediction (IP) \cite{peters2016causal} and LiNGAM \cite{shimizu2006linear} (since there is no undirected edges in the output of IP and LiNGAM, the UD ratio of both would be zero). For LiNGAM, we combined the data from two environments as the input. Therefore, the distribution of the exogenous noise of variables in $I_{12}$ is not Guassian anymore.  As it can be seen in Figure \ref{fig:sim}(a), the error ratio of IP increases as the size of $I_{12}$ increases. This is mainly due to the fact that  in IP approach it is assumed that the distribution of exogenous noise of the target variable should not change, which may be violated by increasing $|I_{12}|$. The result of simulations shows that the error ratio of LiNGAM is approximately twice of those of LRE and PC.


\textbf{Real Data.}
We considered dataset of educational attainment of teenagers \cite{rouse1995democratization}. The dataset was collected from 4739 pupils from about 1100 US high school with 13 attributes including gender, race, base year composite test score, family income, whether the parent attended college, and county unemployment rate. We split the dataset into two parts where the first part includes data from all pupils who live closer than 10 miles to some 4-year college. In our experiment, we tried to identify the potential causes that influence the years of education the pupils received. We ran LRE algorithm on the two parts of data as two environments with a significance level of 0.01 and obtained the following attributes as a possible set of parents of the target variable: base year composite test score, whether father was a college graduate, race, and whether school was in urban area. The IP method \cite{peters2016causal} also showed that the first two attributes have significant effects on the target variable. 


\newpage

\bibliographystyle{abbrv}
\bibliography{Refs}

\newpage

\begin{appendices}

\section{Derivation of Equation \eqref{eq:Ftest}} 
\label{app:eq4}

The null hypothesis $H_{0,Y,S}^{ij}$ can be written in the following form: $\mathbf{C}[\beta_S^{(i)}(Y);\beta_S^{(j)}(Y)]=0$ where $\mathbf{C}$ is a $|S|\times (2|S|)$ matrix such that nonzero entries of $\mathbf{C}$ are $[\mathbf{C}]_{k,k}=1$, $[\mathbf{C}]_{k,k+|S|}=-1$, for all $1\leq k\leq |S|$. Thus, the following statistic 
\begin{equation}
(\hat{\beta}^{(i)}_S(Y)-\hat{\beta}^{(j)}_S(Y))^T (\mathbf{C}\hat{\Sigma}\mathbf{C}^T)^{-1} (\hat{\beta}^{(i)}_S(Y)-\hat{\beta}^{(j)}_S(Y))/p
\end{equation}
has a $F(p,n-p)$ distribution \cite{lutkepohl2005new} where $\hat{\Sigma}=[s_i^2 \Sigma_i^{-1},\mathbf{0}_{|S|\times |S|};\mathbf{0}_{|S|\times |S|},s_j^2 \Sigma_j^{-1}]$. Since $\mathbf{C}\hat{\Sigma}\mathbf{C}^T=s_i^2 \Sigma_i^{-1}+s_j^2 \Sigma_j^{-1}$, the statistic in \eqref{eq:Ftest} has the same $F(p,n-p)$ distribution.

\section{Proof of Lemma \ref{lem:change}}
\label{app:lemchange}

For any set $S\subseteq N(Y)$ and $(Y,S)\in R_{ij}$, using representation \eqref{eq:noise}, we have:
\begin{align*}
& Y^{(i)}=\sum_{X_k\in \textit{AN}(Y)\backslash \{Y\}} c_k N^{(i)}_k +N^{(i)}_Y,\\
& X_{S}^{(i)}\beta_{S}^{(i)}(Y)=\sum_{X_k \in \textit{AN}(Y)\backslash \{Y\}} b_k N^{(i)}_k+\sum_{X_k \in \textit{AN}(S_{CH})\backslash \textit{AN}(Y)} b'_k N^{(i)}_k + b_Y N^{(i)}_Y,
\end{align*}
where $S_{\textit{CH}}:=S\cap \textit{CH}(Y)$ and the ancestral set $AN(X)$ of a variable $X$ consists of $X$ and all the ancestors of nodes in $X$. Moreover, coefficients $b_k$'s and $c_k$'s are functions of $\mathbf{B}$ and $\beta_S(Y)$ which are fixed in two environments. Therefore
\begin{equation}
Y^{(i)}-X^{(i)}_{S}\beta_{S}^{(i)}(Y) = \sum_{X_k \in \textit{AN}(Y)\backslash \{Y\}} (c_k-b_k) N^{(i)}_k -\sum_{X_k \in \textit{AN}(S_{CH})\backslash \textit{AN}(Y)} b'_{k} N^{(i)}_{k} +(1-b_Y) N^{(i)}_Y, 
\label{eq:lemma1}
\end{equation}
If the variance of $N_Y$ is not changed, then clearly for the choice of $S=\textit{PA}(Y)$, the second summation vanishes, and in the first summation $c_k=b_k$. Therefore, the variance of residual remains unvaried. Otherwise, if the variance of $N_Y$ varies, then its change may cancel out only for specific values of the variances of other exogenous noises which according to a similar reasoning as the one in Assumption \ref{ass:stability}, we ignore it.

\section{Proof of Lemma \ref{lem:fv}}
\label{app:lemfv}
Suppose $X$ is the parent of $Y$. Consider environments $E_i, E_j\in\mathcal{E}$.
It suffices to show that if $\beta_Y^{(i)}(X)=\beta_Y^{(j)}(X)$, then $\beta_X^{(i)}(Y)=\beta_X^{(j)}(Y)$.
Using representation \eqref{eq:noise}, $X$ and $Y$ can be expressed as follows
\begin{align*}
&X=\sum_{X_k\in \textit{AN}(X)}a_kN_k\\
&Y=\sum_{X_k\in \textit{AN}(X)}b_kN_k+\sum_{X_k\in \textit{AN}(Y)\backslash\textit{AN}(X)}c_kN_k.
\end{align*}
Hence we have 
\begin{align*}
&\mathbb{E}[X^2]=\sum_{X_k\in \textit{AN}(X)}a^2_k var(N_k)\\
&\mathbb{E}[Y^2]=\sum_{X_k\in \textit{AN}(X)}b^2_kvar(N_k)+\sum_{X_k\in \textit{AN}(Y)\backslash\textit{AN}(X)}c^2_kvar(N_k)\\
&\mathbb{E}[XY]=\sum_{X_k\in \textit{AN}(X)}a_kb_k var(N_k)
\end{align*}
Therefore
\begin{align*}
&\beta_X(Y)=\frac{\sum_{X_k\in \textit{AN}(X)}a_kb_k var(N_k)}{\sum_{X_k\in \textit{AN}(X)}a^2_k var(N_k)}\\
&\beta_Y(X)=\frac{\sum_{X_k\in \textit{AN}(X)}a_kb_k var(N_k)}{\sum_{X_k\in \textit{AN}(X)}b^2_kvar(N_k)+\sum_{X_k\in \textit{AN}(Y)\backslash\textit{AN}(X)}c^2_kvar(N_k)}
\end{align*}
in the expression for $\beta_Y(X)$, the first summation contains the same exogenous noises as the numerator while the second summation contains terms related to the variance of other orthogonal exogenous noises. Therefore, by Assumption \ref{ass:stability}, $\beta_Y^{(i)}(X)=\beta_Y^{(j)}(X)$ only if for all $X_k\in \textit{AN}(Y)$, $var(N_k)$ remains unchanged. In this case, we will also have $\beta_X^{(i)}(Y)=\beta_X^{(j)}(Y)$. Note that $\beta_X(Y)$ can always remain unchanged of the exogenous noise of variables in $\textit{AN}(X)$ affect $Y$ only through $X$.

\end{appendices}

\end{document}